\documentclass[conference]{IEEEtran}
\IEEEoverridecommandlockouts
\usepackage{cite}
\usepackage{amsmath,amssymb,amsfonts}
\usepackage{algorithmic}
\usepackage{graphicx}
\usepackage{textcomp}
\usepackage{xcolor}
\def\BibTeX{{\rm B\kern-.05em{\sc i\kern-.025em b}\kern-.08em
    T\kern-.1667em\lower.7ex\hbox{E}\kern-.125emX}}

\begin{document}

\title{
Improve Load Forecasting in Energy Communities through Transfer Learning using Open-Access Synthetic Profiles
\thanks{The authors gratefully acknowledge the financial support from the Austrian Research Promotion Agency FFG for the Hub4FlECs project (COIN FFG 898053).
\\\\
\copyright~2024 IEEE. Personal use of this material is permitted. Permission from IEEE must be obtained for all other uses, in any current or future media, including reprinting/republishing this material for advertising or promotional purposes, creating new collective works, for resale or redistribution to servers or lists, or reuse of any copyrighted component of this work in other works.
}}

\author{
\IEEEauthorblockA{
    \begin{tabular}{c}
        Lukas Moosbrugger, 
        Valentin Seiler, 
        Gerhard Huber, and 
        Peter Kepplinger\\
        \textit{Research Center Energy, illwerke vkw Endowed Professorship for Energy Efficiency,}\\
        \textit{Vorarlberg University of Applied Sciences, Dornbirn, Austria}\\
        \{molu, seva, huge, kepe\}@fhv.at
    \end{tabular}
}}

\maketitle

%%%%%%%%%%%%%%%%%%%%%%%%%%%%%%%%%%%%%%%%%%%%%%%%%%%%%%%%%%%%%%%%%%%%%%%%%%%%%%%%%%%%%%%%%%
\begin{abstract}
%%%%%%%%%%%%%%%%%%%%%%%%%%%%%%%%%%%%%%%%%%%%%%%%%%%%%%%%%%%%%%%%%%%%%%%%%%%%%%%%%%%%%%%%%%

According to a conservative estimate, a 1\% reduction in forecast error for a 10 GW energy utility can save up to \$ 1.6 million annually. % from: \cite{pinheiro_short-term_2023}
In our context, achieving precise forecasts of future power consumption is crucial for operating flexible energy assets using model predictive control approaches.
Specifically, this work focuses on the load profile forecast of a first-year energy community with the common practical challenge of limited historical data availability. We propose to pre-train the load prediction models with open-access synthetic load profiles using transfer learning techniques to tackle this challenge. 
Results show that this approach improves both, the training stability and prediction error. In a test case with 74 households, the prediction mean squared error (MSE) decreased from 0.34 to 0.13, showing transfer learning based on synthetic load profiles to be a viable approach to compensate for a lack of historic data.

\end{abstract}


%%%%%%%%%%%%%%%%%%%%%%%%%%%%%%%%%%%%%%%%%%%%%%%%%%%%%%%%%%%%%%%%%%%%%%%%%%%%%%%%%%%%%%%%%%
\section{Introduction}
%%%%%%%%%%%%%%%%%%%%%%%%%%%%%%%%%%%%%%%%%%%%%%%%%%%%%%%%%%%%%%%%%%%%%%%%%%%%%%%%%%%%%%%%%%
For the management of energy resources, model predictive control methods are used with great success, as described in \cite{mariano2021review}, \cite{srithapon2023predictive}, and \cite{wohlgenannt2022demand}.
Many of these control approaches react dynamically based on predictions. Thus, forecasting short-term load is crucial for the optimized operation of flexible energy assets \cite{wazirali2023state}. However, considering real applications, the systems often lack enough historical data at the time of commission. This potentially leads to poor system performance. 

The issue described is also faced by energy communities, which are groups of individuals or organizations that come together to produce, manage, and consume energy locally, often with the goal of increasing sustainability and reducing energy costs \cite{european_commission_clean_2019}.
Here the problem of having insufficient load measurements occurs not only during commissioning but also later, namely during the integration of new participants into the energy community.

Several machine learning techniques have been proposed for short-term load forecasting, covering horizons ranging from one hour to one week \cite{pinheiro_short-term_2023}.
Many practical applications use simple auto-regressive models \cite{pinheiro_short-term_2023}. 
According to Kong et al. \cite{kong_short-term_2019} and Muzaffar and Afshari \cite{muzaffar_short-term_2019}, Neural Networks using Long Short-Term Memory (LSTM) are the best suited models for short-term load forecasting. Input features of their forecasting algorithms are date/time information (e.g., clock time, day of the week, holiday), weather features (e.g., ambient temperature, windspeed), and lagged load (e.g., the power consumed on the day before).

The training process of these models can be significantly enhanced through the application of transfer learning, which has emerged as a promising approach for many energy forecasting problems.
This approach is based on the idea that knowledge learned during one task can be re-used to boost the performance on a task related \cite{ribeiro_transfer_2018}. 
In their study, Ribeiro et al. \cite{ribeiro_transfer_2018} focused on the application of transfer learning to predict the energy demand of school buildings. The dataset was modified during pre- and post-processing and only a total of four buildings have been considered. 
Jung et al. \cite{jung_monthly_2020} conducted a comprehensive study, gathering monthly load data spanning fourteen years across 25 districts in Seoul, each subdivided into five categories. Their approach involved pre-training individual models based on the closest profile matches from different other districts. While innovative, this method presents a drawback compared to ours, as it necessitates historical power measurements of the target for accurate comparison.
Xu and Meng \cite{xu_hybrid_2020} manipulate the raw profile, to avoid negative transfer learning, which occurs when knowledge or experience from a source task hinders the performance of a target task, leading to decreased accuracy or effectiveness in learning.
Their approach involves decomposing the input load profile into distinct components -- \textit{irregular}, \textit{trend}, and \textit{seasonal} -- and using transfer learning exclusively for predicting the \textit{irregular} component. In contrast, our method achieves success through end-to-end machine learning, allowing our model to directly process the raw, unmodified load profiles.
Lee and Rhee \cite{lee_individualized_2021} devised a method where individual short-term load forecast models were pre-trained using historical load data from all other households in their study. However, a significant limitation of this approach arises from the privacy sensitivity of load profiles, making it challenging to share such profiles across different energy communities.

According to Pinheiro et al. \cite{pinheiro_short-term_2023}, many forecast models proposed in the literature are evaluated using specific datasets, often selecting particular time intervals and households. This limits the usefulness of the results and models. The same limitations arise when selecting over-specialized data for pre-training.
Therefore, we propose to use synthetic load profiles for pre-training, an end-to-end machine learning model, and characteristic load profiles for training and testing.
In summary, this paper makes the following key contributions:
\begin{enumerate}
    \item By using synthetic load profiles, we eliminate the need to select specific, appropriate load profiles for pre-training. These synthetic load profiles are available open source from the distribution system operator and therefore a much more general approach compared to the approaches in literature.
    \item We use an end-to-end machine learning model, where the load profiles can be fed into the model, eliminating the need for separate preprocessing steps.
    \item We demonstrate the effectiveness of our proposed method using an example energy community, utilizing open-source load data to validate our approach.
\end{enumerate}

%%%%%%%%%%%%%%%%%%%%%%%%%%%%%%%%%%%%%%%%%%%%%%%%%%%%%%%%%%%%%%%%%%%%%%%%%%%%%%%%%%%%%%%%%%
\section{Methods}\label{Methods}
%%%%%%%%%%%%%%%%%%%%%%%%%%%%%%%%%%%%%%%%%%%%%%%%%%%%%%%%%%%%%%%%%%%%%%%%%%%%%%%%%%%%%%%%%%

The experiments are conducted using two datasets and a bidirectional LSTM model, incorporating features typical for load forecasting. Additionally, we consider the real-world limitation of limited training data availability within an energy community. This involves starting with a small amount of training data and receiving new training data on a (only) weekly basis, gradually increasing the dataset over time.

\subsection{Datasets}
Two different load profile sources are used in this study:
\begin{enumerate}
    \item To test the continuous learning of a first-year energy community along with transfer learning, we aggregated the characteristic load profiles of $74$ German households \cite{tjaden_reprasentative_nodate}. The smart meter data collected for the entire year of 2010 were originally recorded at 15-minute intervals, which we down-sampled to hourly intervals. The authors elaborated in \cite{tjaden_reprasentative_nodate} that the load profiles are characteristic in aggregate but further noted that there is no detailed metadata available about these specific loads.
    \label{Dataset1}
    \item The Austrian Power Clearing and Settlement Agency provides synthetic load data \cite{noauthor_synthetic_nodate}. These Austrian synthetic load profiles are used for pre-training prior to training with the German load dataset. They are available some years ahead, which is important for applications such as the one considered here.\label{Dataset2}

\end{enumerate}

The historic weather data are fetched with an hourly resolution using the python library \textit{meteostat}, from a weather station in Bochum (for the load profile dataset \ref{Dataset1}). The weather features considered are described in more detail in Sec. \ref{Features}. For the synthetic profile (dataset \ref{Dataset2}) no weather data are available.

\subsection{Model}

Recurrent neural networks (RNNs) are a widely favored option for sequence-to-sequence tasks. RNNs reuse neural network cells between timesteps and therefore have a much lower weight count compared to conventional feed-forward neural networks. 
A significant advancement in early RNNs was the introduction of memory cells, known as Long Short-Term Memory (LSTM) cells, which enable neural networks to effectively retain information over long sequences \cite{hochreiter_long_1997}.

\begin{figure}[tbph]
\includegraphics[width=0.49\textwidth]{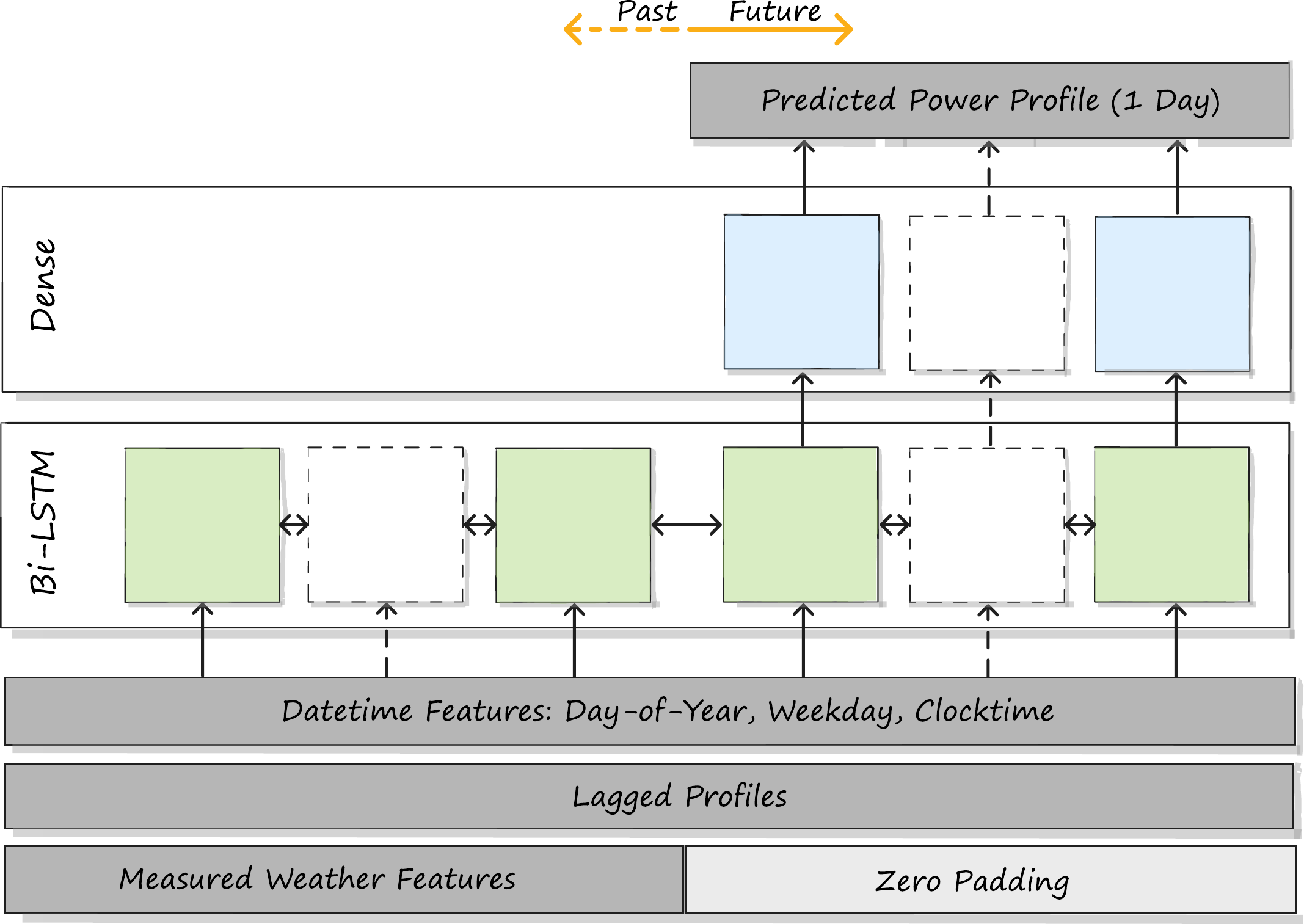}
\caption{Model architecture. 
The model processes an input sequence consisting of 48 hourly time steps and produces an output sequence of 24 hourly time steps, representing the predicted load profile for the following day.
}
\label{fig1}
\end{figure}

The model used is sketched in Figure \ref{fig1}, simplified to only show one bidirectional LSTM (Bi-LSTM) layer of two.
One Bi-LSTM layer consists of two normal LSTM layers with opposing information flow. One passes the extracted information from past to future and the other one passes it from future to the past. 
This bidirectional architecture allows the model to effectively capture both past dependencies and future context within the input data.

The same hyperparameters are used for both pre-training and fine-tuning. Specifically, the Adam optimizer is run for $300$ epochs with a dynamically decreasing learning rate, starting at $0.015$ and gradually reducing to $0.001$. The final learning rate is chosen to match the Keras default, thus avoiding potential overfitting with this hyperparameter. The initial learning rate is selected to accelerate the optimization process, reducing the overall training time to a reasonable duration.
Mean squared error (MSE) is the training loss targeted for optimization.
And finally, both the input and output variables undergo standardization.

In detail, the Keras Sequential model is configured as follows:
\begin{itemize}
    \item Input Layer with shape (48, 18)
    \item Bidirectional LSTM Layer (10 units) with sequence return
    \item Batch Normalization Layer
    \item Bidirectional LSTM Layer (30 units) with sequence return
    \item Batch Normalization Layer
    \item Lambda Layer for selecting the last 24 time steps
    \item Dense Layer (10 units, ReLU activation)
    \item Dense Layer (10 units, ReLU activation)
    \item Dense Layer (1 unit, Linear activation)
    \item Output Layer with shape (24, 1)
\end{itemize}

\subsection{Features}\label{Features}

As reported in \cite{pinheiro_short-term_2023}, the load is largely dependent on human behavior, which in turn is conditioned by the calendar, suggesting that date and time features should be taken into account. Fortunately, date/time features are easy to integrate into real applications.

Furthermore \cite{pinheiro_short-term_2023} states that load forecasts can benefit from weather data. According to this, we can assume that weather data, for instance the outside temperature, strongly affects the total power consumption used for (electrified) heating, whereas the sun influences e.g. the need for artificial lighting. 
We further assume that all energy communities have convenient access to recent local weather measurements.

Another very common feature for load forecasts is lagged load, especially for simple models like auto-regressive algorithms \cite{pinheiro_short-term_2023}. Lagged load refers to using historical load data from previous time steps as input features for the model. For example, the next Sunday could be predicted just by averaging the last few Sundays. Nevertheless, it has to be kept in mind that, depending on the lag time, power measurement data might not be available in time.

In detail, the inputs for the LSTM model are as follows:
\begin{itemize}
    \item \textbf{Inputs 0-6:} Day-of-week, encoded as one-hot, i.e. each day represented by a binary 7-vector with one 'hot' bit. Main public holidays are treated equally to Sundays.
    \item \textbf{Inputs 7-8:} Hour-of-day, cyclically encoded, i.e., the hour-of-day is mapped onto a circle using sine and cosine functions, which preserves the cyclical nature of the data.
    \item \textbf{Inputs 9-10:} Day-of-year, cyclically encoded. The day-of-year is mapped onto a circle using sine and cosine functions.
    \item \textbf{Inputs 11-13:} Lagged load profile. The load measurements from 1, 2, and 3 weeks ago are fed into the model.
    \item \textbf{Inputs 14-17:} Weather data. This feature depends on the used load profile dataset:
    \begin{itemize}
        \item Synthetic load data: These data are synthetic and contain no corresponding weather information. Therefore this feature is set to a vector containing only zeros.
        \item Characteristic German dataset: All available meteostat weather features of Bochum from 2010 are used. This includes "temperature", "precipitation", "wind speed" and "sunshine duration".
    \end{itemize}
\end{itemize}

\subsection{Application to energy communities}\label{method_rec}

\begin{figure}[tbph]
% \begin{figure*}[tbh]
\centering
\includegraphics[width=0.49\textwidth]{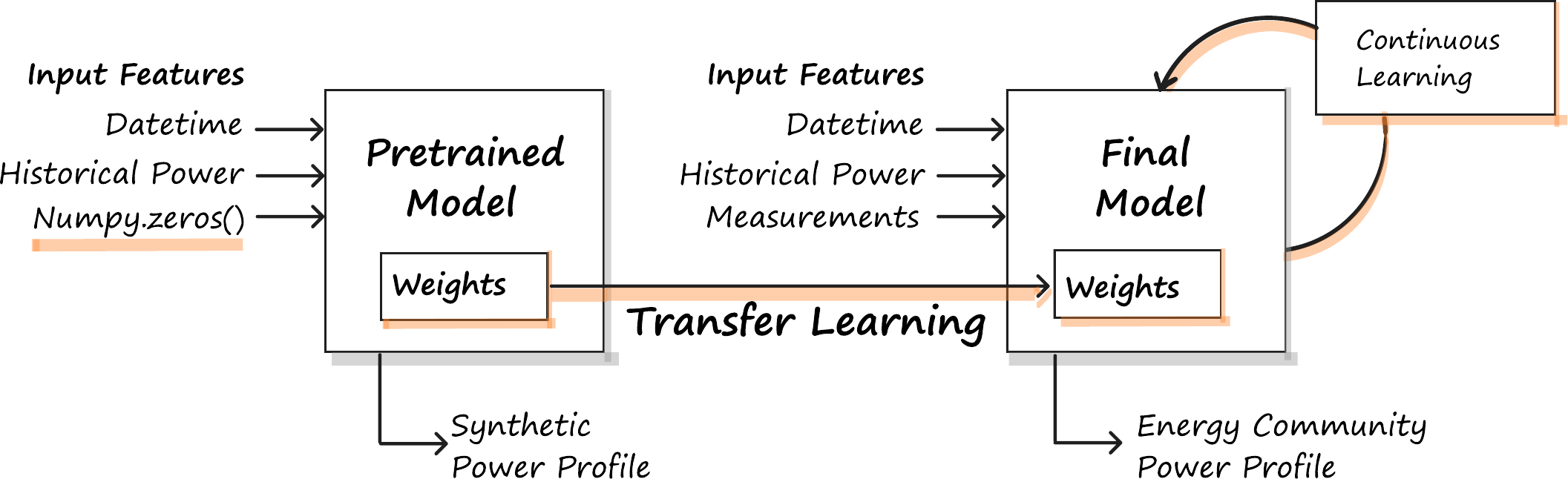}
\caption{Main concept, key ideas are highlighted in orange. 
The model is pre-trained with synthetic load profiles. As no weather information is known from the synthetic load profiles, a new feature will be added during finetuning. The model can start predicting from scratch at the very first deployment and is further trained with every new measurement available, improving steadily.
}
\label{fig3}
% \end{figure*}
\end{figure}

Figure \ref{fig3} illustrates the main concept of this paper, highlighting key ideas in orange: The learning transfer from synthetic load profiles, the adding of features during fine-tuning, and continuous learning.

The following workflow is conducted, and is also applicable to newly established energy communities:
\begin{enumerate}
    \item First the model is trained on the synthetic profile. In our work, the synthetic profile is chosen over the year 2010. The resulting pre-trained weights are stored.
    \item On day 1 the energy community is set up and starts collecting data.
    \item The first training and prediction with the model could be done on day 3. But as our input feature "lagged load" has a lag of 21 days, we start on day 24. 
    \item On day 24, the model is initialized with the pre-trained weights and trained with the measured load profile. With the resulting model, the next 24 hours are predicted and evaluated.
    \item The model is retrained every 7 days on an increasing training set and makes predictions of the next 24 hours every day.
\end{enumerate}

\begin{figure}[tbph]
\includegraphics[width=0.49\textwidth]{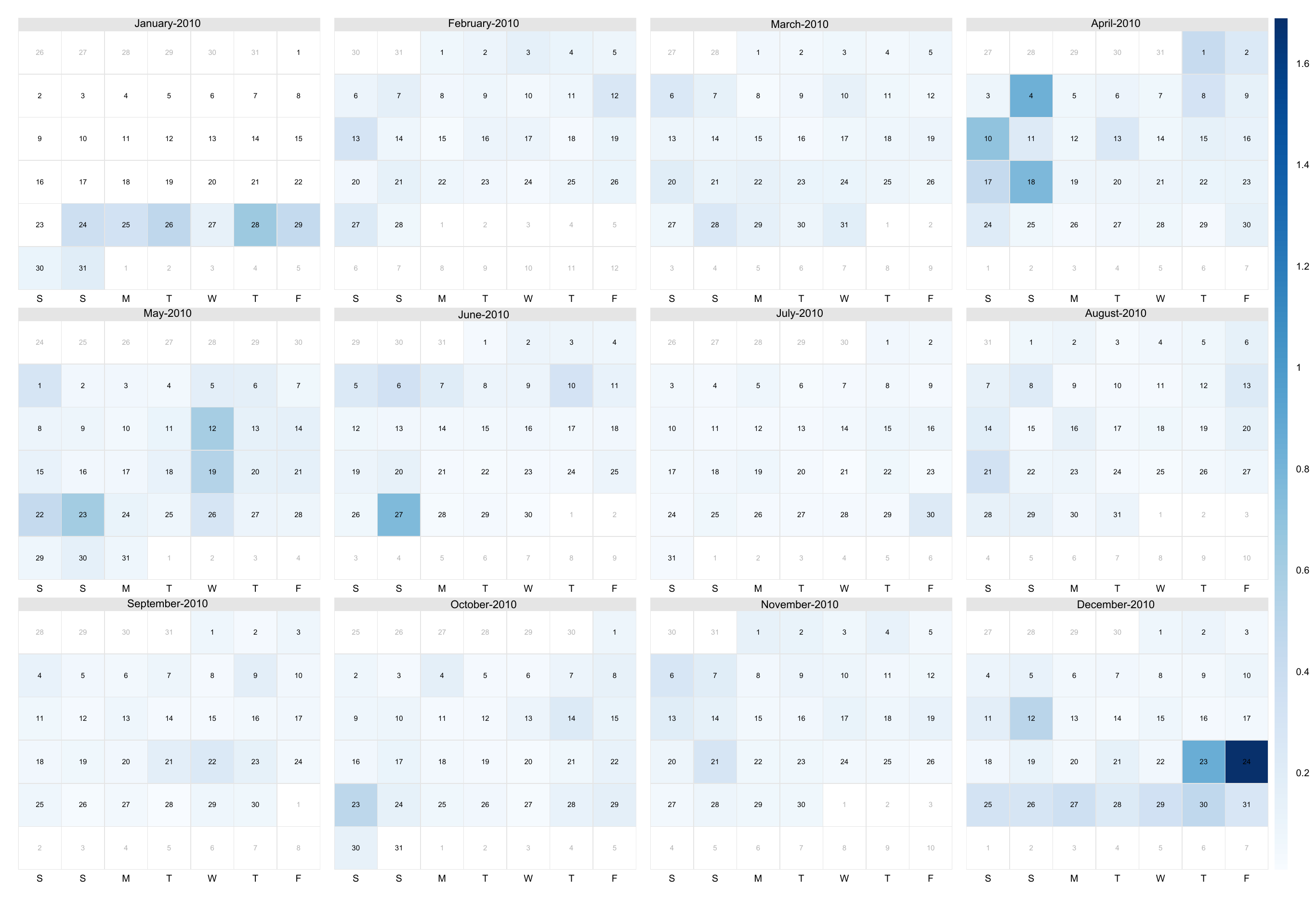}
\caption{The prediction error (MSE) of the continuously trained model over the year 2010.}
\label{fig_calendar_plot}
\end{figure}

%%%%%%%%%%%%%%%%%%%%%%%%%%%%%%%%%%%%%%%%%%%%%%%%%%%%%%%%%%%%%%%%%%%%%%%%%%%%%%%%%%%%%%%%%%
\section{Results}
%%%%%%%%%%%%%%%%%%%%%%%%%%%%%%%%%%%%%%%%%%%%%%%%%%%%%%%%%%%%%%%%%%%%%%%%%%%%%%%%%%%%%%%%%%

%This section first shows the impact of transfer learning with synthetic load data.

Figure \ref{fig_calendar_plot} shows the prediction error of our model (with transfer learning) for each day of the year. 
The procedure applied follows the above description: For a given day, all preceding days are included in the training set, and the trained model is then used to predict the given day.

The imperfect prediction days in figure  \ref{fig_calendar_plot} come quite as expected:
\begin{itemize}
    \item Due to the lagged load profile feature (described in section~\ref{Features}), the first prediction day is not January 3\textsuperscript{rd}, but January 24\textsuperscript{th}. The first week after January 24\textsuperscript{th} shows high prediction errors, because of a very small training set.
    \item Other erroneous days are mostly around public German holidays, e.g. Easter Sunday on April 4\textsuperscript{th}, the evening before Ascension Day on May 13\textsuperscript{th}, Whitsunday on May 23\textsuperscript{rd} or Christmas Eve on December 24\textsuperscript{th}. 
\end{itemize}

\begin{figure}[tbph]
\includegraphics[width=0.5\textwidth]{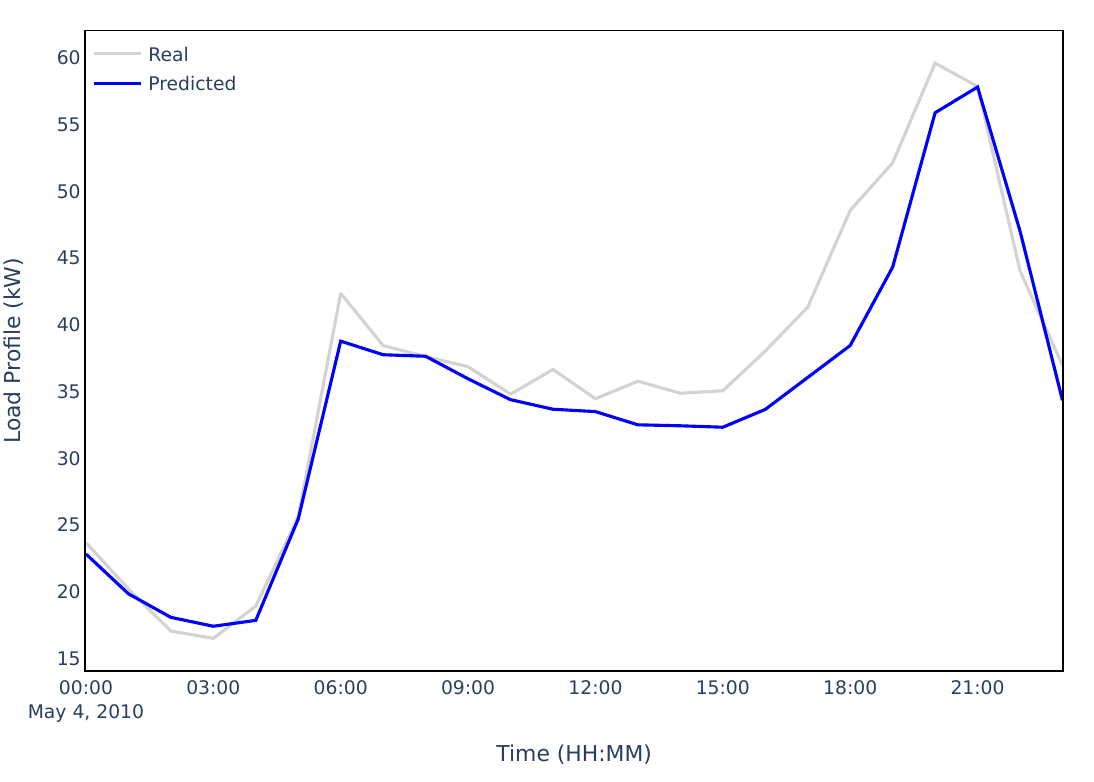}
\caption{Example day with a decent prediction error.}
\label{fig12}
\end{figure}

\begin{figure}[tbph]
\includegraphics[width=0.5\textwidth]{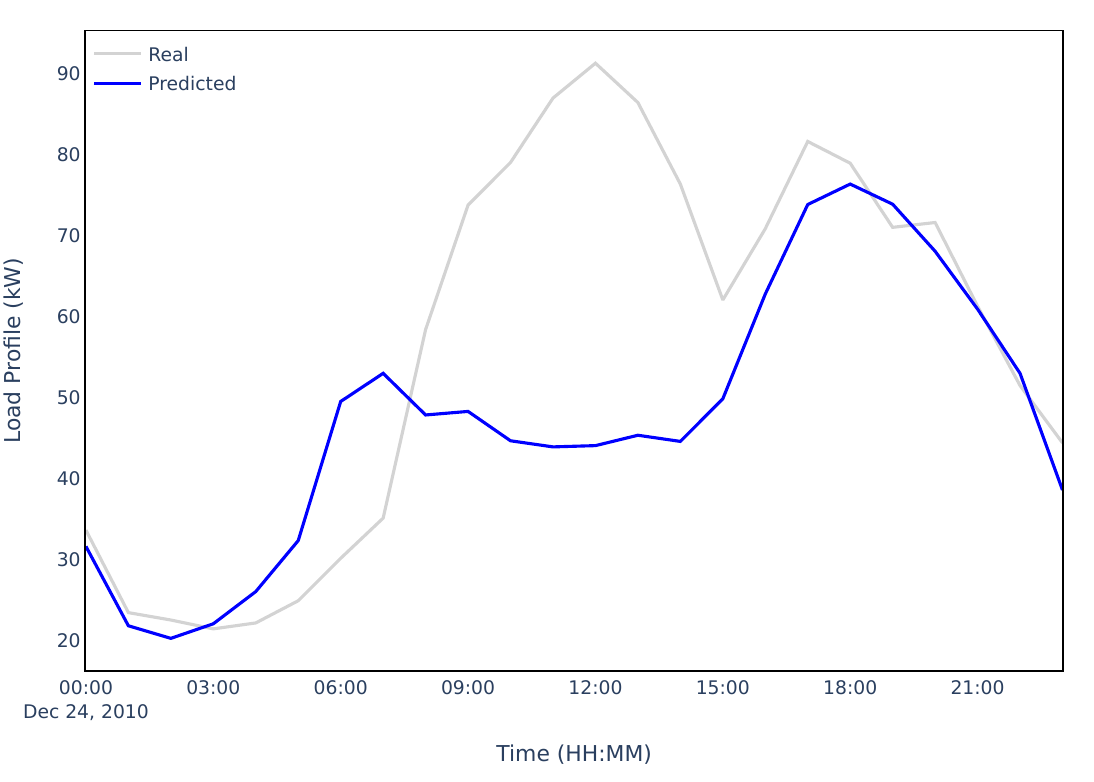}
\caption{December 24\textsuperscript{th} (Christmas Eve) shows the highest prediction error of the whole year.}
\label{fig15}
\end{figure}

Figures \ref{fig12} and \ref{fig15} depict two example days. Figure~\ref{fig12} displays the actual and predicted profiles for May 4\textsuperscript{th}, a day characterized by accurate prediction. Especially the timings of the maxima were predicted well. The second figure shows the load profile of December 24\textsuperscript{th}. Here, especially the load peak during lunchtime was underestimated significantly.

\begin{figure}[tbph]
\includegraphics[width=0.49\textwidth]{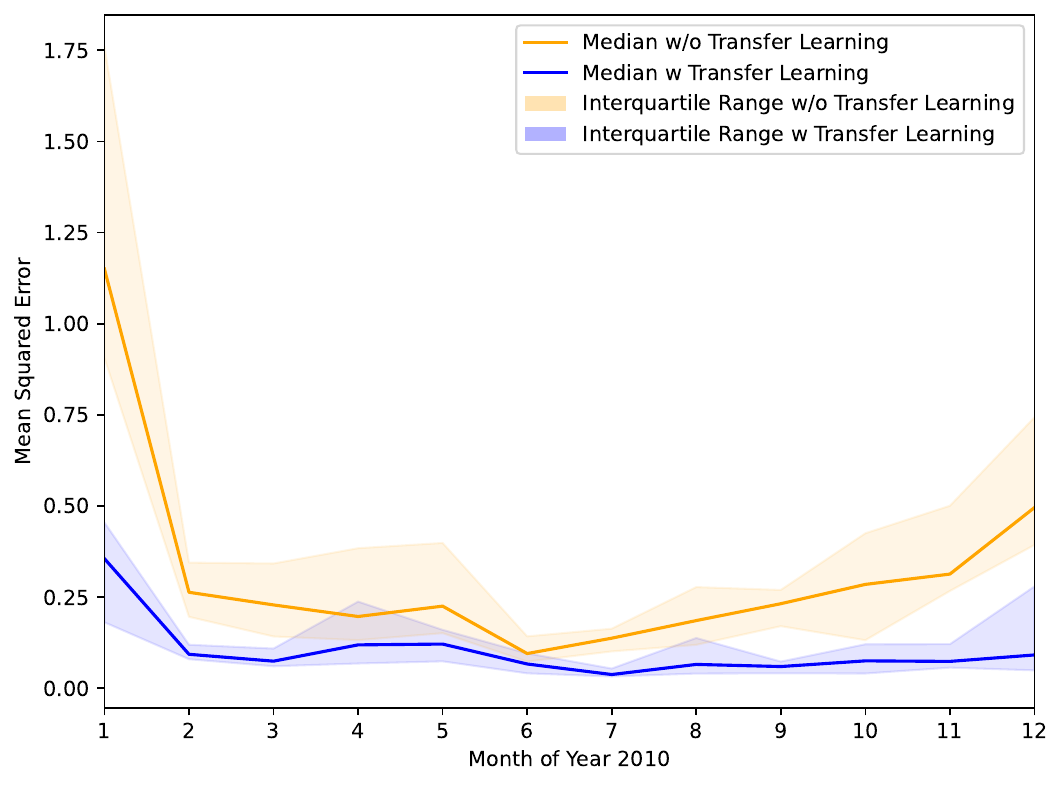}
\caption{Prediction errors of a newly commissioned energy community. Comparison with and without transfer learning from synthetic load profiles.}
\label{fig_transfer_learning_deviation}
\end{figure}

Figure \ref{fig_transfer_learning_deviation} depicts both the monthly prediction errors and prediction deviation. The model that employs transfer learning consistently outperforms the other one. The main reasons for the high error in January and in December are again the initially small training set and the Christmas holidays, respectively.

\begin{figure}[tbph]
\includegraphics[width=0.49\textwidth]{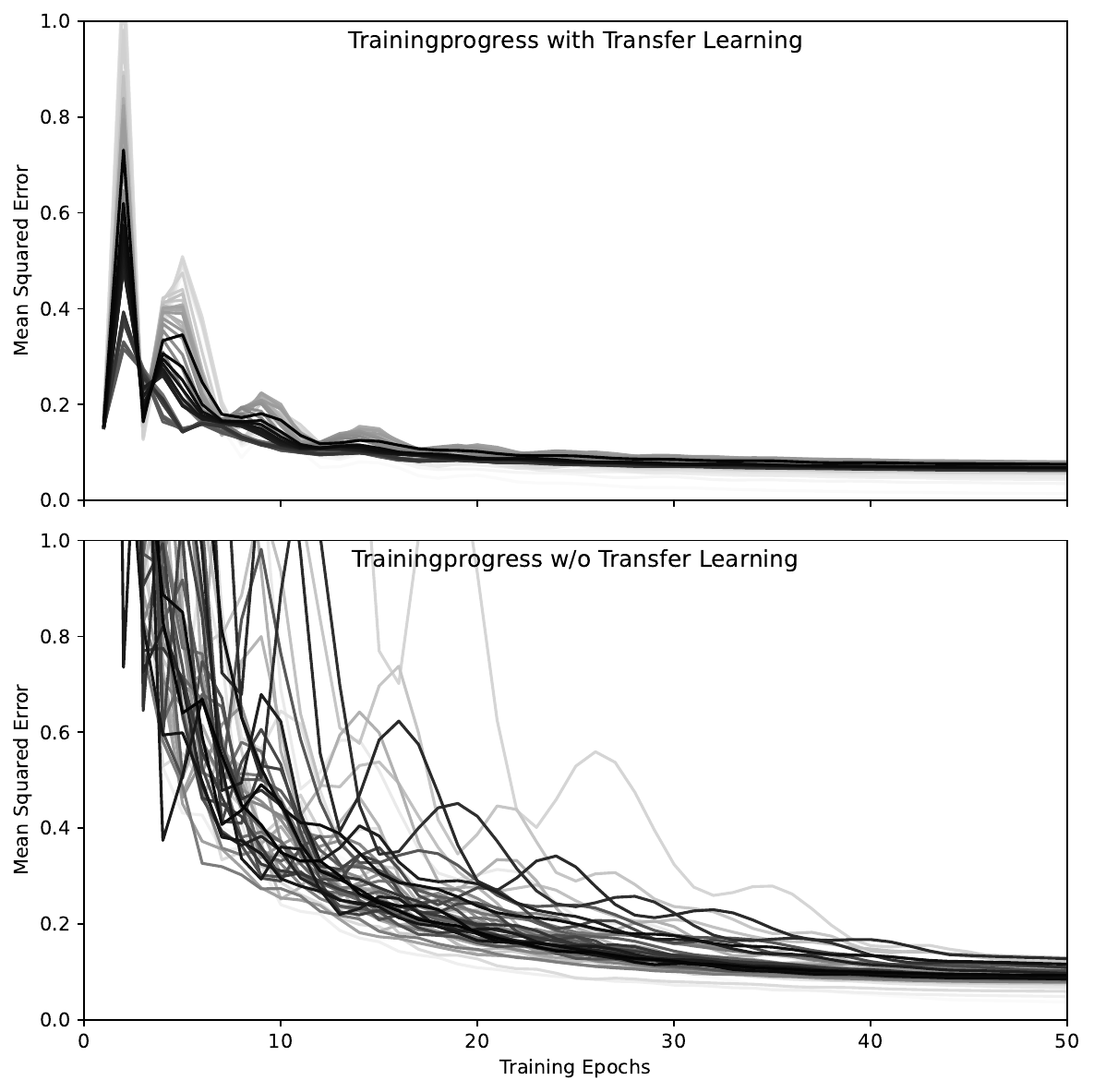}
\caption{Comparison of the training progress of the first 50 epochs. The trace colors transition from light grey (representing the first training day of 2010) to dark grey (representing the last training day of 2010).}
\label{fig_training_loss}
\end{figure}

Besides the prediction error, the training stability increases due to transfer learning with synthetic load profiles, as can be seen in Figure \ref{fig_training_loss}.
Especially the light and medium grey traces in the lower sub-figure, which represent the first weeks after commissioning a new energy community, are only slowly converging.
Moreover, our experiments show that this enhanced training stability is even more beneficial for deeper models. Models with more LSTM units and more dense layers, often exhibit no convergence at all for small datasets without transfer learning with synthetic load profiles.

%%%%%%%%%%%%%%%%%%%%%%%%%%%%%%%%%%%%%%%%%%%%%%%%%%%%%%%%%%%%%%%%%%%%%%%%%%%%%%%%%%%%%%%%%%
\section{Conclusion and Future Work}
%%%%%%%%%%%%%%%%%%%%%%%%%%%%%%%%%%%%%%%%%%%%%%%%%%%%%%%%%%%%%%%%%%%%%%%%%%%%%%%%%%%%%%%%%%

This study proposes an LSTM-based approach to address the short-term load forecasting problem of energy communities. Transfer learning using synthetic load profiles is investigated as a means to overcome the practical barrier of limited historical data.

Specifically, this work demonstrates, that transfer learning from synthetic load profiles brings substantial benefits for both the training process and load prediction of energy communities. The average prediction MSE decreased from $0.34$ to $0.13$ due to transfer learning.

Furthermore, this study opened many avenues for further exploration. While it focused on medium-sized energy communities, extending the research to smaller communities or individual households could be promising. Additionally, analyzing different synthetic load profiles and evaluating Transformer models might prove valuable. The results obtained from Transformers, particularly when using transfer learning, could then be compared to those achieved with LSTMs.

%%%%%%%%%%%%%%%%%%%%%%%%%%%%%%%%%%%%%%%%%%%%%%%%%%%%%%%%%%%%%%%%%%%%%%%%%%%%%%%%%%%%%%%%%%
\bibliographystyle{IEEEtran}
\bibliography{IEEE_RTSI_2024}

\begin{thebibliography}{10}
\providecommand{\url}[1]{#1}
\csname url@samestyle\endcsname
\providecommand{\newblock}{\relax}
\providecommand{\bibinfo}[2]{#2}
\providecommand{\BIBentrySTDinterwordspacing}{\spaceskip=0pt\relax}
\providecommand{\BIBentryALTinterwordstretchfactor}{4}
\providecommand{\BIBentryALTinterwordspacing}{\spaceskip=\fontdimen2\font plus
\BIBentryALTinterwordstretchfactor\fontdimen3\font minus \fontdimen4\font\relax}
\providecommand{\BIBforeignlanguage}[2]{{%
\expandafter\ifx\csname l@#1\endcsname\relax
\typeout{** WARNING: IEEEtran.bst: No hyphenation pattern has been}%
\typeout{** loaded for the language `#1'. Using the pattern for}%
\typeout{** the default language instead.}%
\else
\language=\csname l@#1\endcsname
\fi
#2}}
\providecommand{\BIBdecl}{\relax}
\BIBdecl

\bibitem{mariano2021review}
D.~Mariano-Hern{\'a}ndez, L.~Hern{\'a}ndez-Callejo, A.~Zorita-Lamadrid, O.~Duque-P{\'e}rez, and F.~S. Garc{\'\i}a, ``A review of strategies for building energy management system: Model predictive control, demand side management, optimization, and fault detect \& diagnosis,'' \emph{Journal of Building Engineering}, vol.~33, p. 101692, 2021.

\bibitem{srithapon2023predictive}
C.~Srithapon and D.~M{\aa}nsson, ``Predictive control and coordination for energy community flexibility with electric vehicles, heat pumps and thermal energy storage,'' \emph{Applied Energy}, vol. 347, p. 121500, 2023.

\bibitem{wohlgenannt2022demand}
P.~Wohlgenannt, M.~Prei{\ss}inger, M.~L. Kolhe, and P.~Kepplinger, ``Demand side management of a battery-supported manufacturing process with on-site generation,'' in \emph{2022 IEEE 7th International Energy Conference (ENERGYCON)}.\hskip 1em plus 0.5em minus 0.4em\relax IEEE, 2022, pp. 1--6.

\bibitem{wazirali2023state}
R.~Wazirali, E.~Yaghoubi, M.~S.~S. Abujazar, R.~Ahmad, and A.~H. Vakili, ``State-of-the-art review on energy and load forecasting in microgrids using artificial neural networks, machine learning, and deep learning techniques,'' \emph{Electric power systems research}, vol. 225, p. 109792, 2023.

\bibitem{european_commission_clean_2019}
\BIBentryALTinterwordspacing
{European Commission}, \emph{\BIBforeignlanguage{eng}{Clean energy for all {Europeans}}}.\hskip 1em plus 0.5em minus 0.4em\relax Publications Office of the European Union, 2019. [Online]. Available: \url{https://data.europa.eu/doi/10.2833/9937}
\BIBentrySTDinterwordspacing

\bibitem{pinheiro_short-term_2023}
M.~G. Pinheiro, S.~C. Madeira, and A.~P. Francisco, ``Short-term electricity load forecasting—a systematic approach from system level to secondary substations,'' \emph{Applied Energy}, vol. 332, p. 120493, 2023.

\bibitem{kong_short-term_2019}
W.~Kong, Z.~Y. Dong, Y.~Jia, D.~J. Hill, Y.~Xu, and Y.~Zhang, ``Short-term residential load forecasting based on {LSTM} recurrent neural network,'' \emph{{IEEE} Transactions on Smart Grid}, vol.~10, no.~1, pp. 841--851, 2019.

\bibitem{muzaffar_short-term_2019}
S.~Muzaffar and A.~Afshari, ``Short-term load forecasts using {LSTM} networks,'' \emph{Energy Procedia}, vol. 158, pp. 2922--2927, 2019.

\bibitem{ribeiro_transfer_2018}
M.~Ribeiro, K.~Grolinger, H.~F. {ElYamany}, W.~A. Higashino, and M.~A. Capretz, ``Transfer learning with seasonal and trend adjustment for cross-building energy forecasting,'' \emph{Energy and Buildings}, vol. 165, pp. 352--363, 2018.

\bibitem{jung_monthly_2020}
S.-M. Jung, S.~Park, S.-W. Jung, and E.~Hwang, ``Monthly electric load forecasting using transfer learning for smart cities,'' \emph{Sustainability}, vol.~12, no.~16, p. 6364, 2020.

\bibitem{xu_hybrid_2020}
X.~Xu and Z.~Meng, ``A hybrid transfer learning model for short-term electric load forecasting,'' \emph{Electrical Engineering}, vol. 102, no.~3, pp. 1371--1381, 2020.

\bibitem{lee_individualized_2021}
E.~Lee and W.~Rhee, ``Individualized short-term electric load forecasting with deep neural network based transfer learning and meta learning,'' \emph{{IEEE} Access}, vol.~9, pp. 15\,413--15\,425, 2021.

\bibitem{tjaden_reprasentative_nodate}
T.~Tjaden, J.~Bergner, J.~Weniger, and J.~Quaschning, ``Repräsentative elektrische {Lastprofile} für {Einfamilienhäuser} in {Deutschland} auf 1-sekündiger {Datenbasis},'' {Hochschule für Technik und Wirtschaft (HTW) Berlin,} {Lizenz}: CC-BY-NC-4.0.

\bibitem{noauthor_synthetic_nodate}
\BIBentryALTinterwordspacing
Synthetic load profiles {APCS} - power clearing \& settlement. Accessed on April 24, 2024. [Online]. Available: \url{https://www.apcs.at/en/clearing/physical-clearing/synthetic-load-profiles}
\BIBentrySTDinterwordspacing

\bibitem{hochreiter_long_1997}
S.~Hochreiter and J.~Schmidhuber, ``Long short-term memory,'' \emph{Neural Computation}, vol.~9, no.~8, pp. 1735--1780, 1997.

\end{thebibliography}
\end{document}